\documentclass[10pt,twocolumn,letterpaper]{article}

\usepackage{cvpr}
\usepackage{times}
\usepackage{epsfig}
\usepackage{graphicx}
\usepackage{amsmath}
\usepackage{amssymb}
\usepackage{booktabs}
\usepackage{subfiles}
\usepackage{diagbox}
\usepackage{dblfloatfix}

\usepackage[pagebackref=true,breaklinks=true,letterpaper=true,colorlinks,bookmarks=false]{hyperref}

\cvprfinalcopy % *** Uncomment this line for the final submission

\def\cvprPaperID{3} % *** Enter the CVPR Paper ID here

\newcolumntype{C}[1]{>{\centering\arraybackslash}p{#1}}

\newcommand{\C}{\mathcal{C}}
\newcommand{\Cbase}{\mathcal{C}_{\text{base}}}
\newcommand{\Cnovel}{\mathcal{C}_{\text{novel}}}
\newcommand{\SData}{\mathcal{S}}
\newcommand{\Strain}{\mathcal{S}_{\text{train}}}

\newcommand{\Strainnovel}{\mathcal{S}_{\text{train}}^{\text{novel}}}
\newcommand{\Strainbase}{\mathcal{S}_{\text{train}}^{\text{base}}}

\ifcvprfinal\fi
\let\svthefootnote\thefootnote
\begin{document}

%%%%%%%%% TITLE
\title{Cross-modal Hallucination for Few-shot Fine-grained Recognition}

\author{Frederik Pahde$^{1,2}$, Patrick J{\"a}hnichen$^2$, ~Tassilo Klein$^1$, ~Moin Nabi$^1$\\
$^1$ Machine Learning Research, SAP SE, Berlin, Germany \\
$^2$ Humboldt Universit{\"a}t zu Berlin \\
\texttt{frederik.pahde@student.hu-berlin.de$^*$, patrick.jaehnichen@hu-berlin.de,}\\ \texttt{\{tassilo.klein, m.nabi\}@sap.com}
}

%\and

\maketitle

\begin{abstract}
State-of-the-art deep learning algorithms generally require large amounts of data for model training. Lack thereof can severely deteriorate the performance, particularly in scenarios with fine-grained boundaries between categories. To this end, we propose a multimodal approach that facilitates bridging the information gap by means of meaningful joint embeddings. Specifically, we present a benchmark that is multimodal during training (i.e. images and texts) and single-modal in testing time (i.e. images), with the associated task to utilize multimodal data in base classes (with many samples), to learn explicit visual classifiers for novel classes (with few samples). Next, we propose a framework built upon the idea of cross-modal data hallucination. In this regard, we introduce a discriminative text-conditional GAN for sample generation with a simple self-paced strategy for sample selection. We show the results of our proposed discriminative hallucinated method for 1-, 2-, and 5-shot learning on the CUB dataset, where the accuracy is improved by employing multimodal data.\\
%Experiments on our proposed benchmark demonstrate that training generative models in a cross-modal fashion facilitates few-shot learning by compensating the lack of data in novel categories, outperforming the single-modal baseline in the challenging 1-, 2-, 5- and 10-shot fine-grained categorization tasks.
% * <engarpe861@gmail.com> 2018-04-28T19:32:36.683Z:
% 
% > we propose a multimodal approach that facilitates bridging the information gap by means of meaningful joint embeddings. Specifically, we present a benchmark that is multimodal during training (i.e. images and texts) and single-modal in testing time (i.e. images), with the associated task to utilize multimodal data in base classes (with many samples), to learn explicit visual classifiers for novel classes (with few samples)
% This sentence is too long
% 
% ^.
% * <engarpe861@gmail.com> 2018-04-28T19:32:00.642Z:
% 
% > To this end
% I would start directly with " We propose"
% 
% ^.
%\textbf{Keywords:} Few-Shot Learning, Multimodal, Meta-Learning, Fine-grained Recognition
\end{abstract}
\let\thefootnote\relax\footnote{$^*$Alternative e-mail address: frederikpahde@gmail.com}
\addtocounter{footnote}{-1}\let\thefootnote\svthefootnote

\begin{figure*}[ht]
	\centering
  \includegraphics[width=.85\textwidth]{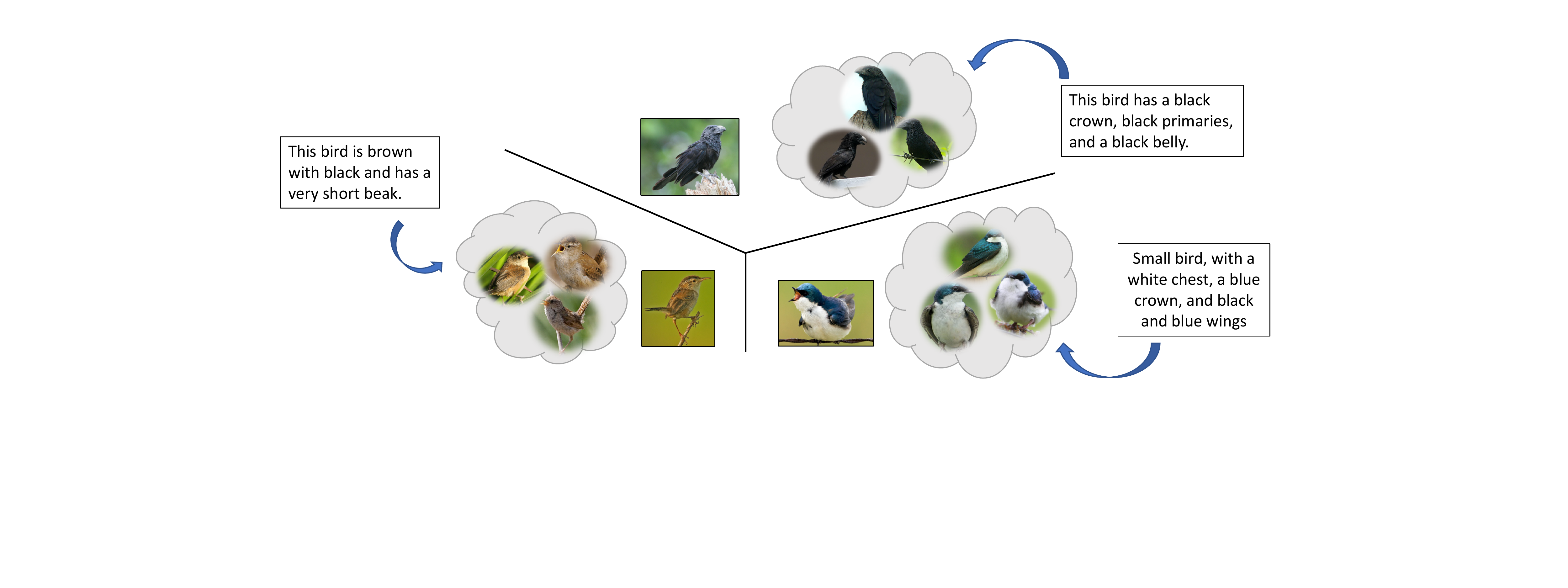}
	\caption{Learn classifier from dataset with a few samples per category extended by hallucinated data conditioned on textual descriptions}
	\label{fig:task}
\end{figure*}
\section{Introduction}
%In recent years, deep learning techniques have achieved remarkable results in many domains such as computer vision and NLP, constantly pushing the boundaries of what is possible. These advances can be explained by improvements to algorithms and model architecture along with increasing computational power and in particular growing availability of big data. However, the big data assumption, which is key for deep learning applications, is at the same time a the limiting factor. For many applications, it is often too expensive or even impossible to acquire sufficiently many training samples in order to learn a model at sufficient accuracy. Furthermore, the requirement for large amounts of training data is in stark contrast to human learning. This is what makes alternative learning approaches that require less training data a very attractive research topic. Thus research in the domain of few-shot learning, i.e. learning and generalizing from few training samples, has gained more and more interest (e.g. \cite{ravi_optimization_2016}, \cite{hariharan_low-shot_2016},
%\cite{snell_prototypical_2017}). \\
%The proposed approaches tackles this problem by grounding salient detail information in a semantically aligned multi-modal embedding space, which is obtained from data-rich base classes. This in turn allows for the cross-modal generation of data that compensates the lack of data of novel categories in the few-shot scenario.

%Recent research topics
%\begin{itemize}
%%\item Multi-modal data:
%\cite{karpathy_deep_2015},
%\cite{faghri_vse++:_2017}
%\end{itemize}
In recent years, deep learning techniques have achieved exceptional results in many domains such as computer vision (e.g. \cite{krizhevsky2012imagenet,szegedy2014googlenet}) and natural language processing (e.g. \cite{2016naacl}). These advances can be explained by improvements to algorithms and model architecture along with increasing computational power, and growing availability of big data. 
The big data assumption is key for conventional deep learning applications but often also a limiting factor. Particularly for fine-grained recognition tasks, the existence of sufficient training samples is necessary \cite{li2017zero}. 
%However, the big data assumption, which is key for deep learning applications, is at the same time the limiting factor. 
However, for many applications, it is often too expensive or even impossible to acquire enough training samples in order to learn a model at sufficient accuracy. 
Furthermore, the requirement for large amounts of training data is in stark contrast to human learning, which can quickly learn from few instances. 
This makes alternative learning approaches that require less training data an attractive research topic. 
For that reason, research in the domain of few-shot learning, i.e. learning and generalizing from few training samples, has gained more and more interest (e.g. \cite{koch_siamese_2015, ravi_optimization_2017, snell_prototypical_2017, vinyals_matching_2016}). Specifically, most of the current works (e.g. \cite{hariharan_low-shot_2017}) have proposed meta-learning-based approaches. These assume the existence of some \emph{base classes} with many training samples that can be used to learn powerful representations which, in turn, can be employed to perform classification on \emph{novel classes} with only a few samples. 
However, research conducted has mainly focused on approaches with data coming from only one modality, primarily images. 
By overcoming the single-modality restriction and including data from additional modalities, limitations in the low data regime can be overcome, resulting in improved model performance. 
The key assumption is that incorporating multimodal data, i.e. images and fine-grained descriptions thereof, forces the model to identify highly discriminative features \textit{across modalities}, facilitating training in few-shot scenarios.
% (e.g. parts in the image and textual attributes in the description).
Specifically, pursuing multimodality suggests that novel concepts with low amount of training data in one modality can benefit from previously learned alignments between the two modalities, such that existing data in the additional modality (e.g. text) can compensate the lack of data in the other modality (e.g. image).
%Overcoming this limitation and including additional data from other modalities, e.g. textual descriptions, can further improve the model. The assumption of our approach is that having fine-grained descriptions provided from multimodal data can force the model to focus on the more discriminative features (e.g. parts and attributes) of novel classes in order to achieve improved performance in the few-shot learning setting \cite{elhoseiny_link_2017}. 
% * <engarpe861@gmail.com> 2018-04-28T19:33:58.489Z:
% 
% > However
% Don't need this conjunction here
% 
% ^.
This assumption leads to the proposed study of few-shot learning with multimodal data, more precisely images with fine-grained textual descriptions. 
The principal contribution of this paper is to extend few-shot learning to deal with multimodal data. Specifically, a scenario is assumed, that is multimodal during training (i.e. images and texts) and single-modal during testing time (i.e. images). Hence, the multimodal data is exploited during training, but the ultimate task remains to train an image classifier.
We address this problem from a cross-modal generative perspective (e.g. \cite{ mishra2017generative, mishra2018generative, sharma2018chatpainter, zhang_stackgan++:_2017}), combining ideas from meta-learning which have been put forward by Hariharan et al.~\cite{hariharan_low-shot_2017} in conjunction with a simple self-paced learning strategy for sample selection. The intuition behind our method is that we generate additional training samples (\cite{sixt2016rendergan, Zhu2017DataAI}) conditioned on textual descriptions that facilitate learning classification models in low data scenarios (see Fig.~\ref{fig:task}).\\ %This is illustrated in Fig~\ref{fig:task}.\\
%\begin{figure*}[ht]
%	\centering
%  \includegraphics[width=.9\textwidth]{figures/taskFigure.jpg}
%	\caption{Learn classifier from dataset with only a few samples per category extended by hallucinated data conditioned on textual descriptions}
%	\label{fig:task}
%\end{figure*}
The most closely related work to the proposed approach is by Hariharan et al.~\cite{hariharan_low-shot_2017} and Wang et al.~\cite{wang_low-shot_2018}, who similarly use hallucinated data for few-shot learning with the difference of the restriction to a single-modal image context. Analogously, Zhang et al.~\cite{zhang_stackgan++:_2017} and Reed et al.~\cite{reed16_gen} proposed to use Generative Adversarial Networks (GANs) \cite{goodfellow_generative_2014,salimans_improved_2016} to generate images from textual descriptions. They, however, just employed it in a zero-shot fashion, ignoring a few number of samples available of novel classes. \\
Our contribution in this work is two-fold: \textbf{First}, we propose a benchmark for multimodal few-shot learning based on the challenging fine-grained visual recognition task that is multimodal during training and single-modal (i.e. images) during test time. \textbf{Second}, we develop a class-discriminative text-conditional generative adversarial network (tcGAN) that facilitates few-shot learning by hallucinating additional images conditioned on fine-grained textual descriptions. Our approach features robustness and outperforms the single-modality baseline in the challenging few-shot scenario on the fine-grained CUB dataset.
% * <engarpe861@gmail.com> 2018-04-28T19:35:47.101Z:
% 
% > we develop a class-discriminative text-conditional generative adversarial network (tcGAN) that facilitates few-shot learning by hallucinating additional images conditioned on fine-grained textual descriptions, featuring robustness and outperforming the single-modality baseline in the challenging low-shot scenario.
% Also a very long sentence
% 
% ^.

\begin{table*}
\caption{Comparison of few-shot learning tasks}
\label{tab:benchmarks}
\begin{center}
\begin{tabular}{lcccc} \toprule
			Task  & Base Classes & Novel Classes  & Multimodal Training Data & Examples\\ \midrule
Classic One-shot Learning 	& no & yes & no	& \cite{fei-fei_one-shot_2006,lake_one_2011,koch_siamese_2015}\\
  Few-shot Learning with Meta-Learning & yes & yes	& no	&	\cite{ravi_optimization_2017,hariharan_low-shot_2017,snell_prototypical_2017}\\
    Multi-modal Few-shot Learning	& yes & yes	& yes	& Ours	\\
 \bottomrule
\end{tabular}
\end{center}
\end{table*}
\section{Related Work}
\textbf{Few-Shot Learning:} For learning with limited amounts of data, Koch et al.~\cite{koch_siamese_2015} proposed a metric learning approach for which siamese convolutional networks were used in a one-shot learning scenario to rank the similarity of inputs. Other work seeks to avoid overfitting by modifications to the loss function or the regularization term. Yoo et al.~\cite{yoo_efficient_2017} proposed a clustering of neurons on each layer of the network and calculated a single gradient for all members of a cluster during the training to prevent overfitting. A more intuitive strategy is to approach few-shot learning on data-level, meaning that the performance of the model can be improved by finding strategies to enlarge the training data. For example, Douze et al.~\cite{douze_low-shot_2017} proposed a semi-supervised approach in which a large unlabeled dataset containing similar images was included in addition to the original training set. Hariharan et al.~\cite{hariharan_low-shot_2017} combined both strategies (data-level and algorithm-level) by defining the squared gradient magnitude (SGM) loss on the one hand and generating new images by hallucinating features on the other hand. Other recent approaches to few-shot learning have leveraged meta-learning strategies. Ravi et al.~\cite{ravi_optimization_2017} trained a long short-term memory (LSTM) network as meta-learner that learns the exact optimization algorithm to train a learner neural network that performs the classification in a few-shot learning setting. Vinyals et al.~\cite{vinyals_matching_2016} introduced matching networks for one-shot learning tasks. This network is able to apply an attention mechanism over embeddings of labeled samples in order to classify unlabeled samples. Snell et al.~\cite{snell_prototypical_2017} proposed prototypical networks which can be interpreted as generalization for matching networks. Prototypical networks search for a non-linear embedding space in which classes can be represented as the mean of all corresponding samples, called a prototype and classification is performed by finding the closest prototype in the embedding.%learned embedding.

\textbf{Multi-modal Learning:}
By defining a encoder-decoder pipeline, Kiros et al.~\cite{kiros_unifying_2014} proposed a method to align visual and semantic information in a joint embedding space.
Faghri et al.~\cite{faghri_vse++:_2017} were able to improve this mixed representation by incorporated a triplet ranking loss. The work of Karpathy et al.~\cite{karpathy_deep_2015} aims to generate image descriptions. Their model is able to infer latent alignments between regions of the image and segments of the sentences for the image description. Reed et al.~\cite{reed_learning_2016} put their focus on fine-grained visual descriptions. They collected two datasets containing fine-grained visual descriptions and proposed a deep structured joint embedding that is end-to-end trainable.

\textbf{Conditional GANs:}
After the introduction of GANs by Goodfellow et al.~\cite{goodfellow_generative_2014}, the conditional generation of data was investigated by Mirza et al.~\cite{mirza_conditional_2014}. Reed et al.~\cite{reed_generative_2016} further studied image synthesis based on textual information. Using multiple stacked GANs on top of each other,~Han et al.~\cite{han2017stackgan} pushed the quality of the generated images to photo-realistic high-resolution level by stacking multiple GANs. Following that work, many improvements have been suggested. While in the optimization of StackGAN~\cite{han2017stackgan} all GANs are trained successively, the same authors have developed an end-to-end trainable version, which they named StackGAN++~\cite{zhang_stackgan++:_2017}. Further improvements include the incorporation of an attention mechanism over the textual input (Xu et al.~\cite{xu_attngan:_2017}) and the usage of additional dialog data (Sharma et al.~\cite{sharma2018chatpainter}).

%\section{Related Work}
%\subfile{sections/relatedWork}

%\begin{figure*}[ht]
%	\centering
 % \includegraphics[width=.75\textwidth]{figures/benchmark.jpg}
%	\caption{Selective image generation with tcGAN}
%	\label{fig:tcGAN}
%\end{figure*}
\section{Multimodal Few-shot Learning Benchmark}
%\begin{table*}[b!]
%\caption{Comparison of few-shot learning tasks}
%\label{tab:benchmarks}
%\begin{center}
%\begin{tabular}{lcccc} \toprule
%			Authors  & Dataset & Amount Classes & Amount Images & Multimodal Data \\ \midrule
% Koch et al.~\cite{koch_siamese_2015} 	& Omniglot \cite{lake2015human} & 1,623	& 32,460	& no\\
%  Vinyals et al.~\cite{vinyals_matching_2016}	& miniImageNet~\cite{imagenet_cvpr09,vinyals_matching_2016} & 100	& 60,000	& no	\\
%    Hariharan et al.~\cite{hariharan_low-shot_2017}	& ImageNet~\cite{imagenet_cvpr09} & 1,000	& 14,197,122	& no	\\
 %  Ours 	& CUB-200 \cite{WahCUB_200_2011}	& 200 & 11,788	& yes \cite{reed_learning_2016}	\\
 %\bottomrule
%\end{tabular}
%\end{center}
%\end{table*}

%\textbf{A Multimodal Few-shot Learning Benchmark: } 
The goal is to build a benchmark for multimodal few-shot fine-grained recognition that mimics situations that arise in practice. Therefore, we propose a few-shot learning benchmark inspired by Hariharan et al.~\cite{hariharan_low-shot_2017} and extend it to work with multimodal training data. Following their work, the idea is to model a few-shot learning framework which consists of multiple phases. The first phase can be used to learn a meaningful representation on a large training set (\emph{representation learning phase}). In a next phase the learned representation can be applied and finetuned for novel categories with few samples (\emph{few-shot learning phase}). This is in contrast to classical one-shot learning settings (e.g. \cite{fei-fei_one-shot_2006, lake_one_2011}) in which no base classes with many samples were available (see Tab.~\ref{tab:benchmarks}).
To this end, let $\mathcal{I}$ denote the image space, $\mathcal{T}$ the text space and $\C=\lbrace 1,...,Y\rbrace$ be the discrete label space. Further, let $x_i \in \mathcal{I}\times \mathcal{T}$ be the $i$-th input data point and $y_i \in \C$ its label.
Following \cite{hariharan_low-shot_2017}, two disjunct subsets of the label space are considered in order to setup a few-shot learning setting: $\Cbase$, that are labels for which a large amount of data samples is available; and novel classes $\Cnovel$ which are underrepresented in the data and just a few instances per category are accessible. Note that both subsets exhaust the label space $\C$, i.e. $\C = \Cbase \cup \Cnovel$. It can further be assumed that in general $|\Cnovel| < |\Cbase|$. This is necessary in order for being able to learn powerful representations.
Furthermore, the data set $\SData$ is organized as followed:
Training data $\Strain$ consists of tuples $\{(x_i,y_i)\}_{i=1}^{n}$ taken from the whole data set with $y_i \in \Cbase \cup \Cnovel$. Hence, the training data is composed of $\Strain = \Strainnovel \cup \Strainbase$, where $\Strainnovel = \{(x_i, y_i) : (x_i, y_i) \in \Strain, y_i \in \Cnovel\}_{i=1}^k \subset \Strain$ and $\Strainbase = \{(x_i, y_i) : (x_i, y_i) \in \Strain, y_i \in \Cbase\}_{i=1}^{n-k} \subset \Strain$. Furthermore, in accordance with a few-shot scenario let 
$
%k = 
\left|\Strainnovel\right|\ll\left|\Strainbase\right| 
%= n_{train}
$. 
Contrary to the benchmark defined by Hariharan et al.~\cite{hariharan_low-shot_2017} and other popular few-shot learning tasks, our scenario is multimodal in training (see Tab.~\ref{tab:benchmarks}). However, the testing phase is single-modal on image data of $\Cnovel$. That means, the classifier is evaluated on image data only to fulfill the ultimate goal to train a visual classifier.

%To this end, we split the classes $\C$ into base classes $\Cbase$ for which many samples exist and novel classes $\Cnovel$ with just a few samples with $\C = \Cbase \cup \Cnovel$ and $\Cbase \cap \Cnovel = \emptyset$. In this scenario, data from $\Cbase$ is used to learn meaningful representations in order to perform few-shot learning on $\Cnovel$. Our proposed benchmark is multimodal in training and single-modal in testing. 
%To that end, the task is to utilize multimodal data of the training set, to learn explicit visual classifiers for novel classes.
%Hence, the training samples are tuples $x_j = \left(I_j,T_j\right)$ consisting of an image $I_j \in \mathcal{I}$ and a textual description $T_j \in \mathcal{T}$, where $\mathcal{I}$ and $\mathcal{T}$ denote the image space and text space, respectively. In test time, however, the model is tested only with the image data from $\Cnovel$. 

\section{Method}
The overall framework of the proposed method can be split into two phases: 1) representation learning in which a discriminative text-conditional GAN is trained to hallucinate images given a textual description and 2) a finetuning phase in which we learn to pick the most discriminative images out of the generated data with a self-paced sample selection strategy. Finally, we train a generic classifier.
\begin{figure*}[ht]
	\centering
  \includegraphics[width=0.85\textwidth]{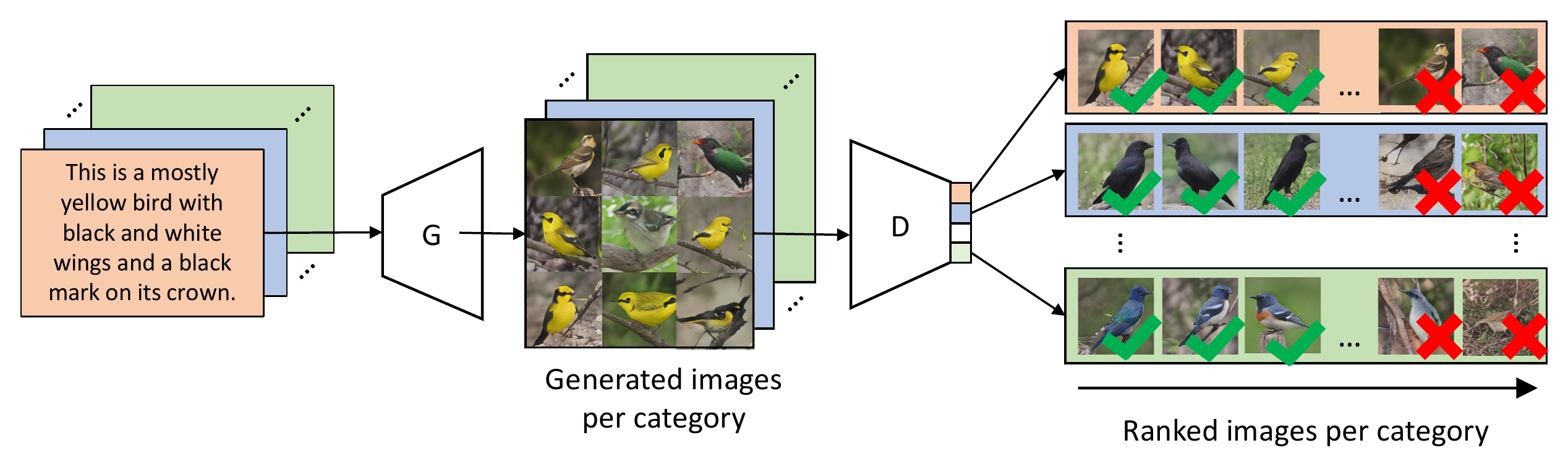}
	\caption{Selective image generation with tcGAN}
	\label{fig:tcGAN}
\end{figure*}
\subsection{Discriminative Text-Conditional GAN} 
Inspired by Wang et al.~\cite{wang_low-shot_2018}, we follow a meta-learning framework and learn a generative model on the large amount of data available in $\Cbase$, then utilize it to learn a classifier for the limited samples related to $\Cnovel$. Therefore, we built a text-conditional GAN (tcGAN) (e.g. \cite{mirza_conditional_2014, reed16_gen, zhang_stackgan++:_2017}) to learn the mapping $\mathcal{T}\rightarrow \mathcal{I}$, such that the generator $G$ is trained to produce outputs that cannot be distinguished from ``real'' images by an adversarially trained discriminator $D$, which is trained to do as well as possible at detecting the generator’s ``fakes''. This allows for cross-modal sample generation, which facilitates few-shot learning by compensating the lack of data in $\Cbase$.\\ 
%With a slight abuse of the notation, the objective of a tcGAN from observed text $T$ and image $I$ can be expressed as:
The objective of a tcGAN from observed text $T$ and image $I$ can concisely be expressed as:
% * <engarpe861@gmail.com> 2018-04-28T19:37:06.095Z:
% 
% > Simplified
% I don't think this is a simplification. This is just to be more concise. But you can omit it and start diretly with "The objective...
% 
% ^.
\begin{multline}
\mathcal{L}_{tcGAN}\left(G,D\right)=\mathbb{E}_{I,T}\left[\log D\left(I,T\right)\right]\\+\mathbb{E}_{I,z}\left[\log D\left(I,G\left(T,z\right)\right)\right],
\end{multline}
where $z$ denotes a random noise vector, and $T$ and $I$ the embeddings for observed text and image respectively.\\
In practice, we built our method on top of the StackGAN framework proposed by Zhang et al.~\cite{zhang_stackgan++:_2017}, which is a variant of tcGAN that features a robust pipeline for generating realistic images from fine-grained textual descriptions. \\
Optimization of the tcGAN loss $\mathcal{L}_{tcGAN}$ alone, however, lacks class-discriminativeness. Therefore we augment $\mathcal{L}_{tcGAN}$ by adding a class-discriminative term $\mathcal{L}_{class}$, which is defined as:
\begin{equation}
\mathcal{L}_{class}\left(D\right)=\mathbb{E}\left[P\left(C=c\mid I\right)\right],
\end{equation}
where $c$ denotes the class label.
Furthermore, let
\begin{equation}
\mathcal{L}_{class}\left(D\right)=\mathcal{L}_{class}\left(G\right).
\end{equation}
This leads to two loss terms:
\begin{equation}\mathcal{L}\left(D\right)=\mathcal{L}_{tcGAN}\left(G,D\right)+\mathcal{L}_{class}\left(D\right)\end{equation}
and
\begin{equation}\mathcal{L}\left(G\right)=\mathcal{L}_{tcGAN}\left(G\right)-\mathcal{L}_{class}\left(G\right),
\end{equation}
which are optimized in alternative fashion, yielding $D^{*}$ and $G^{*}$. It should be noted that whereas $\mathcal{L}_{tcGAN}$ is trained on samples from $\Cbase$, the compound loss is trained only on the (sub-)set of $n$ training samples that are available within $\Cnovel$. This provides us with the training of the tcGAN in a meta-learning fashion, where the cross-modal representation learned on the base classes, is later employed for the final class-discriminative few-shot learning task.

\subsection{Self-paced Sample Selection}
Training the text-conditioned GAN allows for the generation of a potentially infinite number of samples given textual descriptions using $G^*$. However, the challenge is to pick adequate samples out of the pool of generated samples that allow for building a better classifier within the fine-grained few-shot scenario. Such a subset of images should not only be realistic but also class-discriminative. To this end, we follow the \emph{self-paced strategy} and select a subset of images corresponding to ones in which the generator is most confident about their ``reality'' and the discriminator is the most confident about their ``class discriminativeness''. Specifically, we use the score computed using $D^*$ per category and sort generated images in a descending order using these scores. Then we select the first $m$ top-most elements. Intuitively, we select a subset of the generated samples that the classifier trained on the real data is most confident about, as illustrated in Fig.~\ref{fig:tcGAN}. Finally, a convolutional neural network (CNN) is trained on the concatenated set of real images and those ones selected as the best generated class-discriminative images. 
% * <engarpe861@gmail.com> 2018-04-28T19:38:55.522Z:
% 
% > To this end
% Not needed
% 
% ^.
% * <engarpe861@gmail.com> 2018-04-28T19:38:28.674Z:
% 
% > allows for the generation
% maybe just "generates" reads easier
% 
% ^.

\section{Experimental Results}
\begin{figure*}[ht]
	\centering
  \includegraphics[width=0.96\textwidth]{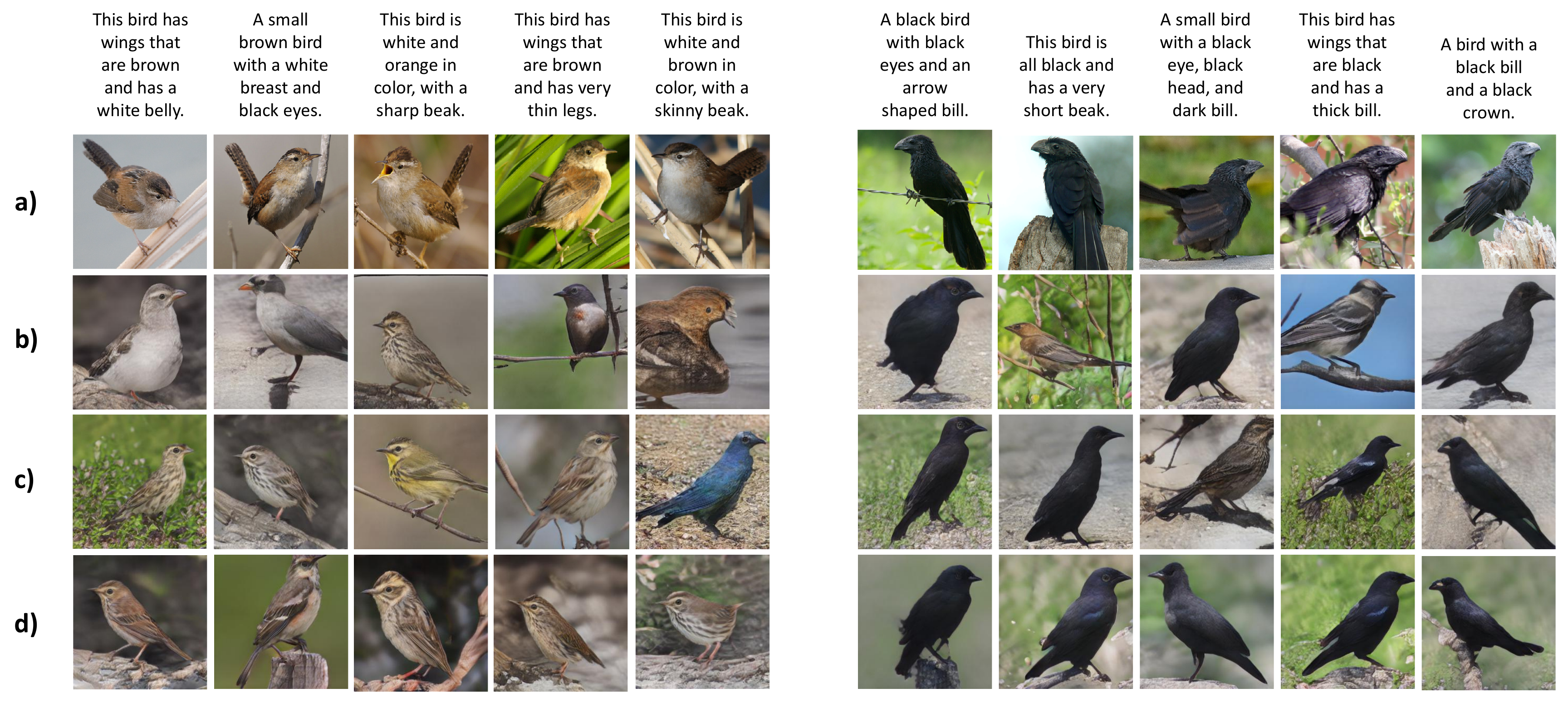}
	\caption{Images for two different bird categories: a) real image data, b) data generated with StackGAN, c) data generated with StackGAN and ranked with $D$, d) data generated with StackGAN and ranked with $D^*$.}
	\label{fig:generatedData}
\end{figure*}

For our experiments we use the CUB dataset~\cite{WahCUB_200_2011}, which contains 11,788 images of 200 different bird species. The data is split equally in training and test data, resulting in roughly 30 training and 30 test images per category. 10 short textual descriptions per image are provided by Reed et al.~\cite{reed_learning_2016}.  Following Zhang et al.~\cite{zhang_stackgan++:_2017} we use a pre-trained text-encoder~\cite{reed_learning_2016} and split the data such that $\left|\Cbase\right|=150$, $\left|\Cnovel\right|=50$. To perform few-shot learning $n=\{1,2,5,10,20\}$ images of $\Cnovel$ are used for training, as proposed by Hariharan et al.~\cite{hariharan_low-shot_2017}. For the sake of simplicity, a CNN with basic architecture is employed for classification, although any other classifier is applicable. It consists of two convolutional layers paired with max-pooling, followed by two linear layers that are connected with dropout completed with a softmax of $|C_{Novel}|$ units. For training SGD is used for 800 epochs with a learning rate of 0.01 and momentum of 0.5. 
The experiments are composed of: 
\textbf{1)} \emph{Single modality baseline (Baseline):} %we train the classifier only on real data, i.e. $n$ images per category.
a baseline is evaluated, in which the classifier is trained exclusively on real samples. Thus, in the $n$-shot scenario, only $n$ images are available per category. %Note that the network is not pre-trained.
\textbf{2)} \emph{Few-shot StackGAN baseline (StackGAN):} 
the generator $G$ is obtained from the representation learning phase in which a StackGAN was trained. Then, $G$ is employed to generate additional training images conditioned on one caption randomly chosen (out of 10) for the missing $m=30-n$ images of $\Cnovel$. Following the notion of image generation conditioned on the chosen descriptions, the classifier is trained on an extended dataset that contains the few real images and the generated samples.
\textbf{3)} \emph{StGD baseline:} To show the importance of our proposed discriminative tcGAN, we generated a large amount of images for captions using the $G$ of StackGAN and then rank these samples by the score of $D$ (real vs. fake discriminative). To this end, samples with a low visual appearance are ranked low. Note that this experiment slightly differs from our method, as we proposed to employ $D^*$ (class-discriminative) instead of $D$ (real vs. fake) for the ranking.
\textbf{4)} \emph{Our proposed method (StGD$^*$): } Similar to the previous baseline with the difference of employing the class-discriminative $D^*$ for ranking generated images. Doing that, images are not picked based on their realistic appearance but on how class-discriminative they are. 
The top-5 accuracy of the classifier for our different experiments is reported in Tab.~\ref{tab:result}.\\ 
% * <engarpe861@gmail.com> 2018-04-28T19:39:27.966Z:
% 
% > accuracy of 
% I would say it is "the top accuracy {values} of the ...."
% 
% ^.
We observe that the proposed approach outperforms the baseline in the particular challenging few-shot scenarios with $n={1,2,5}$ by 4.9 to 8.6 percentage points, respectively. Additionally to commonly reported top-5 accuracy, we evaluated the experiments with top-1 and top-3 accuracy, observing a similar performance results. Using the score of $D$ as measure to rank the generated images alone has shown not to be sufficient. However, enforcing class-discriminativeness within the discriminator leads to significantly higher accuracies.
Our experiments confirm that multimodality allows to close the information gap in few-shot scenarios, yielding more robust classifiers.\\
%Visualizing the data generated in our experiments shows the same observation (see figure \ref{fig:generatedData}), as the birds picked by $D'$ have the most class-discriminative features.
%\hspace{0.01cm}
\begin{table}[h]
%\caption{Top-5 accuracy in percent of our classifier on only real data (R), generated data on random captions (StG), generated data assessed with $D$ (StGD) and generated data assessed with $D'$ (StGD'). Best results are bold.}
%\caption{Top-5 accuracy in percent of our classifier for different setups (R, StG, StGD, StGD$^*$)}
\caption{Top-5 accuracy of our classifier for different training setups}
\label{tab:result}
\begin{center}
\begin{tabular}{lccccc} \toprule
			& & & n	& 	&\\
			Method  & 1 & 2 & 5 & 10 & 20\\ \midrule
 \emph{Baseline} 	& 19.9	& 24.8 & 36.8	& 49.2	& \textbf{70.6}	\\
  \emph{StackGAN}~\cite{zhang_stackgan++:_2017}	& 27.3	& 30.7 & 37.4	& 45.3	& 69.0	\\
   \emph{Our (StGD)} 	& 25.7	& 30.1 & 36.8	& 50.5	& 68.6	\\
   \emph{Our ( StGD$^*$)}	&  \textbf{28.5}	& \textbf{31.6}& \textbf{41.7}	& \textbf{52.2
   }	& 68.5	\\
 \bottomrule
\end{tabular}
\end{center}
\end{table}
Further, qualitative analysis by means of visualization of generated data (see Fig.~\ref{fig:generatedData}) in our experiments confirms that images ranked high by $D^*$ contain the most class-discriminative features (d). In contrast to that, only picking random descriptions as input for the tcGAN leads to an undesirable large variety of birds because many descriptions do not include sufficient class-discriminative information (b). Further, ranking on $D$ produces realistic looking images, however, mixing categories (c).

\section{Conclusion}
In this paper, we proposed to extend few-shot learning for fine-grained recognition to deal with multimodal data and introduced a discriminative tcGAN for sample generation along with a self-paced strategy for sample selection. Experiments on our proposed benchmark demonstrate that learning generative models in a cross-modal fashion facilitates few-shot learning for fine-grained categories by compensating the lack of data in novel categories. %Our results on our proposed benchmark suggest that learning generative models in a cross-modal fashion compensate the lack of data in novel categories%, so facilitating the few-shot learning. 
For future work we plan to investigate the use of $D^*$ as the final classifier. Furthermore, we seek to incorporate class-discriminativeness property at the representation learning stage jointly with the ranking loss of the self-paced stage. %This can for example be achieved by adding a pairwise ranking loss for $D$ during the training of the StackGAN. 
%Last, future research will be conducted on optimizing the embedding in context of multi-modality.

%\textcolor{red}{Please: 1. add more references}\textcolor{green}{ 2. add names of authors (me as last)}\textcolor{green}{3. Use more of the term of ``meta-data'' in the intro/abstract/conclusion}\textcolor{green}{ 4. rename the name of the baseline R to \emph{baseline} and the StG as \emph{StackGAN} 5. }

{\small
\bibliographystyle{ieee}
\bibliography{egbib}
}

\end{document}